\documentclass[10pt,twocolumn,letterpaper]{article}

\usepackage{ijcb}
\usepackage{times}
\usepackage{epsfig}
\usepackage{graphicx}
\usepackage{amsmath}
\usepackage{amssymb}
\usepackage{multirow}
\usepackage{array}
\newcolumntype{L}{>{\centering\arraybackslash}m{0.6\linewidth}}
\usepackage{soul}

\usepackage[pagebackref=true,breaklinks=true,letterpaper=true,colorlinks,bookmarks=false]{hyperref}

\ijcbfinalcopy % *** Uncomment this line for the final submission

\ifijcbfinal\fi
\begin{document}

%%%%%%%%% TITLE
\title{FDeblur-GAN: Fingerprint Deblurring using Generative Adversarial Network}

\author{Amol S. Joshi\hspace{0.5cm} 
%Institution1 address\\
%{\tt\small asj00003@mix.wvu.edu}
% For a paper whose authors are all at the same institution,
% omit the following lines up until the closing ``}''.
% Additional authors and addresses can be added with ``\and'',
% just like the second author.
% To save space, use either the email address or home page, not both

Ali Dabouei\hspace{0.5cm}
%West Virginia University\\
%First line of institution2 address\\
%{\tt\small ad0046@mix.wvu.edu}
Jeremy Dawson\hspace{0.5cm}
%West Virginia University\\
%First line of institution2 address\\
%{\tt\small jeremy.dawson@mail.wvu.edu}
Nasser M. Nasrabadi\vspace{0.2cm}\\
West Virginia University\\
%First line of institution2 address\\
%{\tt\small nasser.nasrabadi@mail.wvu.edu}
{\tt\small \{asj00003, ad0046\}@mix.wvu.edu},
{\tt\small \{jeremy.dawson, nasser.nasrabadi\}@mail.wvu.edu}
}

\maketitle
\thispagestyle{empty}

%%%%%%%%% ABSTRACT
\begin{abstract}
   While working with fingerprint images acquired from crime scenes, mobile cameras, or low-quality sensors, it becomes difficult for automated identification systems to verify the identity due to image blur and distortion. We propose a fingerprint deblurring model FDeblur-GAN, based on the conditional Generative Adversarial Networks (cGANs) and multi-stage framework of the stack GAN. Additionally, we integrate two auxiliary sub-networks into the model for the deblurring task. The first sub-network is a ridge extractor model. It is added to generate ridge maps to ensure that fingerprint information and minutiae are preserved in the deblurring process and prevent the model from generating erroneous minutiae. The second sub-network is a verifier that helps the generator to preserve the ID information during the generation process. Using a database of blurred fingerprints and corresponding ridge maps, the deep network learns to deblur from the input blurry samples. We evaluate the proposed method in combination with two different fingerprint matching algorithms. We achieved an accuracy of 95.18\% on our fingerprint database for the task of matching deblurred and ground truth fingerprints.
\end{abstract}

%%%%%%%%% BODY TEXT
\section{Introduction}
Identifying humans with unique biometric features is a great advantage to security and verification tasks. It offers immense benefits in cases such as crime-solving, evidence collection, and human verification and authorization. Recent work in the field of biometric identification, along with advanced algorithms in deep learning, have demonstrated the reliability of such systems for identification tasks. Even though different biometric systems are adopted in the real world, some traits, like the iris or the face, have their own limitations. Hence, under many circumstances, fingerprints are certainly ideal for identification scenarios. The ease of collection of fingerprints using sensors, devices, mobile cameras, ink, etc. is a unique advantage in fingerprint biometrics. Based on the collection techniques, the fingerprints could be contact-based, i.e., collected using a sensor, or contact-less, such as captured with a camera or a mobile phone. With the advent of robust commercial fingerprint recognition systems, these samples are easy to recognize and verify. 
\begin{figure}[t]
\begin{center}
\includegraphics[width=0.9\linewidth]{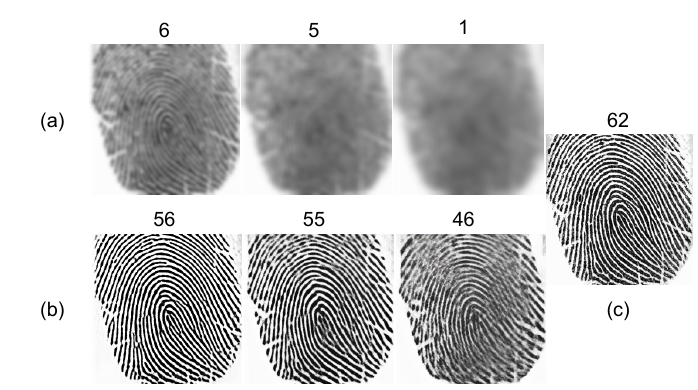}
\end{center}
   \caption{Blurred and respective clean samples with different $\sigma$ values. Row (a) contains blurred samples with $\sigma=3$, $5$, and $7$, respectively, row (b) contains deblurred version of the same fingerprint, and column (c) contains ground truth fingerprint. The quality scores depicted above the samples demonstrate the quality of the fingerprints computed using NFIQ v2.0.}
\label{fig:deblurred_samples}
\end{figure}

However, latent fingerprints acquired from crime scenes or samples captured by the law enforcement agents using the camera of a cellphone may not be as clean as desired. Two types of distortions often deteriorate the quality of fingerprints: geometric distortion and photometric distortion. The geometric distortion, caused by the elasticity of the human skin, degrades the identification information by modifying the relative properties of minutiae in fingerprints. On the other hand, the photometric distortion is mainly caused by the non-ideal conditions of the capturing device, i.e., out-of-focus lenses, perspective distortion, dirt, or moisture on the skin. Blurring is a common type of photometric distortion that can be caused by several factors, such as human errors, trembling fingers, the slow frame rate of the capturing sensor \cite{5773529}, inappropriate focusing of the camera, or intentional blurring by malicious users to evade being identified, etc. Recognition of such blurred samples is a cumbersome task which even state-of-the-art systems may fail to accomplish. 

In this paper, we propose a deep convolutional neural network model, FDeblur-GAN, to deblur the fingerprint samples. We use a conditional generative adversarial network (cGAN) as the core model for the deblurring task and enhance its functionality by considering three modifications. In the first modification, we extract the intermediate features from different layers of the generator and feed them to their corresponding discriminators, which allows the network to train on the low-resolution fingerprint images. In the second modification, we use a deep fingerprint verifier to force the generator model to preserve the ID of the deblurred fingerprints during the deblurring process. In the third modification, we use a deep fingerprint ridge extractor to compute the ridge maps of the deblurred fingerprints. Unlike other image deblurring tasks, in fingerprint deblurring, the ridge information is important. The deblurring process may cause loss of crucial information from the fingerprints and add redundant ridge-valley patterns or non-existing minutiae to the deblurred fingerprint. Hence, we use a pre-trained ridge extraction model to ensure the generator is preserving the minutiae while deblurring fingerprints. All three networks, the discriminators, ridge extractor, and verifier, work simultaneously to enhance the deblurring process. Some sample results are shown in Figure \ref{fig:deblurred_samples}. In nutshell, our contributions are as follows:

\begin{itemize}
    \item We incorporate deep generative models to address the challenging but less explored issue of ridge blurring in fingerprints. 
    \item Two auxiliary models including a deep fingerprint verifier and ridge extractor are incorporated to force the main generator to preserve the ID and ridge information of fingerprints during the deblurring. 
    \item The performance of deblurring is further enhanced by designing a multi-stage generative process based on a coarse-to-fine supervision.  
\end{itemize}

\begin{figure*}
\begin{center}
%\fbox{\rule{0pt}{2in} \rule{.9\linewidth}{0pt}}
\includegraphics[width=0.75\linewidth]{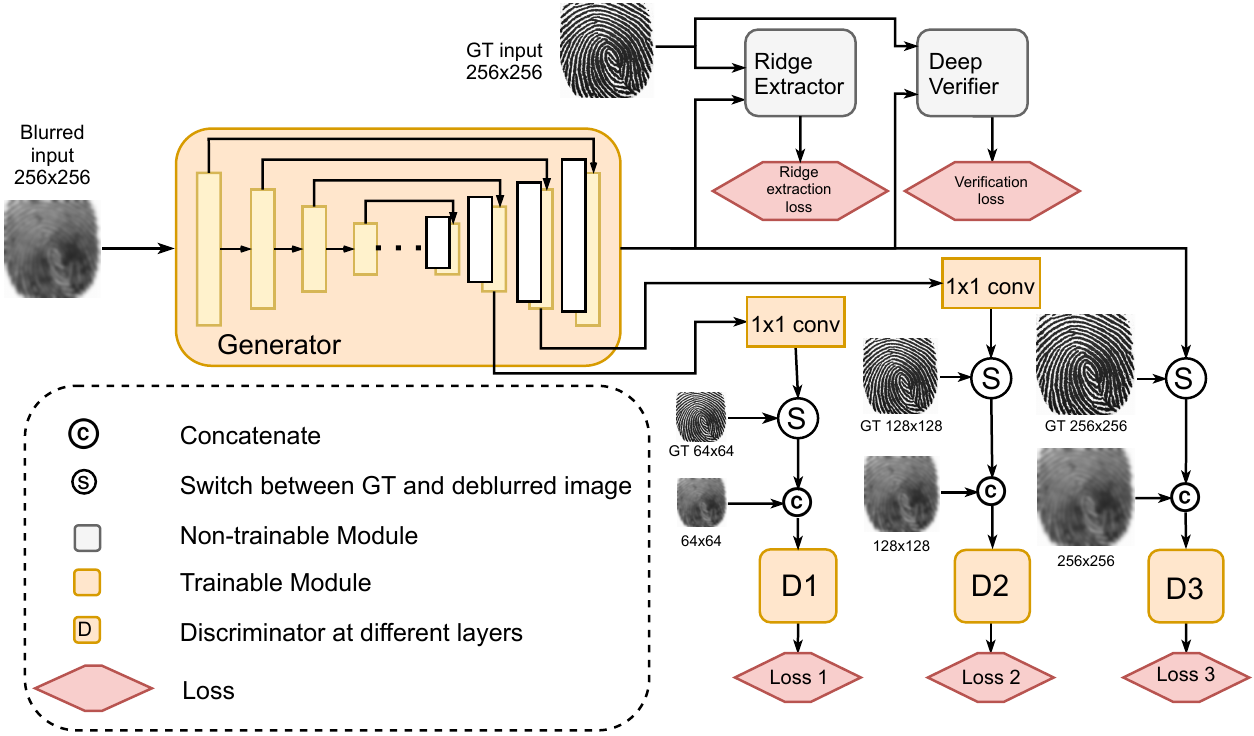}
\end{center}
   \caption{Architecture of the proposed FDeblur-GAN, where the ridge extractor and verifier are the two sub-networks. Input to the generator is a blurred 256x256 image and intermediate outputs are $64\times64$ and $128\times128$. Output of the generator is a $256\times256$ deblurred image. Each intermediate feature along-with the last output have separate discriminators D1, D2, and D3, respectively.}
\label{fig:network}
\end{figure*}

%------------------------------------------------------------------------
\section{Related Work}
Automated fingerprint identification systems consist of three major parts including preprocessing, feature extraction, and matching. In the literature, several techniques have been explored to preprocess fingerprint images, enabling them to serve the intended purpose. The authors in \cite{article} presented an algorithm for enhancing low-quality fingerprints using two processes: alignment and fingerprint enhancement. However, the results of the algorithm depend on the segmentation performance, which may not work as desired on blurred fingerprints. In \cite{GREENBERG2002227}, the authors proposed two techniques for enhancing the fingerprint images. The first technique is based on local histogram equalization, Wiener filtering, and image binarization, whereas the second technique uses a unique anisotropic filter for direct grayscale enhancement. The preprocessing step leads to feature extraction, which is crucial in identification tasks. Kaur \textit{et al.} \cite{5508569} developed a novel smoothing algorithm to extract minutiae using eight different masks. Many studies have been conducted to improve feature extraction \cite{RATHA19951657}. Deductively, the performance of current algorithms for feature extraction and matching of fingerprints depends highly on the quality of the input samples. In \cite{doi:10.1080/19393555.2017.1383535}, the authors studied the effects of low-quality distorted or noisy fingerprint images on recognition performance. In \cite{Chaudhari2013ASA}, they thoroughly illustrated the importance of quality fingerprints for identification. 

\noindent{\bf Photometric distortion correction:} One of the most significant problems in fingerprint biometric is distortions. Based on the cause of the distortion, multiple approaches have been considered with a goal of improving recognition performance. In \cite{8494089}, they implemented distortion detection and rectification algorithm based on geometric features such as orientation maps. For classification of the distortion, they used a feed-forward neural network. Dabouei \textit{et al.} \cite{8411196} used a deep convolutional neural network to rectify the photometric distortion to improve the recognition performance.

\noindent{\bf Blurring in fingerprints:} While working with low-quality fingerprints, such as blurred or distorted images, additional preprocessing steps, such as deblurring, restoration, reconstruction, etc., are inevitable. The deep learning literature presents numerous algorithms and techniques to undertake image reconstruction and restoration. Deep models are now widely used in reconstruction or recognition tasks \cite{WONG2020107203}. The study in \cite{8698580} has developed a certain technique using orientation, frequency, and ridge maps along with a perceptual ID preserving approach which improved the reconstruction of partial latent fingerprints. Svoboda \textit{et al.} \cite{8272727} used a GAN-based approach to predict the missing region of the fingerprint from minutiae and ridge maps. In \cite{6928426}, the authors used another learning approach to reconstruct partial fingerprints. However, to the best of our knowledge, the fingerprint deblurring problem has not been addressed adequately.

\noindent{\bf Deblurring for natural images:} Deblurring can be classified into two types: blind and non-blind. In blind deblurring the kernel of the blurring effect is unknown. There are various techniques for blind deblurring \cite{9008540} as well as non-blind deblurring \cite{6618928}. Several attempts have been made in the literature to deblur images with different causes of blurring, such as motion blur, Gaussian blur, etc. For instance, \cite{9008540} used a pyramid network as a generator with a GAN-based model to remove motion blur from natural images. Using a double scale discriminator and light-weight backbones, they improved deblurring accuracy and efficiency. Tao \textit{et al.} \cite{Tao_2018_CVPR} implemented a scale-recurrent neural network to deblur single images. They used a coarse-to-fine strategy such that, at every scale, a sharp latent image is produced. 

Be that as it may, existing methods for deblurring of biometric images are mostly focused on other biometric traits such as iris \cite{5206700}, face \cite{alaoui}, and hand-based biometric traits such as palm prints \cite{7126364}  or finger wrinkles \cite{9035411}.  Cho \textit{et al.} \cite{9035411} proposed a GAN-based model for deblurring finger wrinkles for authentication. Having mentioned these techniques, fingerprint deblurring has not been explored enough. Even though deep image deblurring models achieve much higher accuracy, a deblurred fingerprint needs ID preservation to be matched with the ground truth fingerprint.
%------------------------------------------------------------------------
\section{Proposed Method}
In this section, we describe our proposed FDeblur-GAN framework consisting of a cGAN deblurring network with two sub-networks arranged together to perform efficient deblurring of fingerprints. 
\subsection{Conditional GAN}\label{sec:cgan}
When it comes to cross-domain transformation, GANs \cite{Goodfellow2014GenerativeAN} are the most popular generative networks. These networks map the input sample from a random distribution $p_{z}(Z)$ to another domain such that $y = G(z,\theta g) : z \xrightarrow{} y$ , where $\theta_{g}$ represents the trained parameters of the network. GANs are usually a pair of two networks, a generator and a discriminator. The role of the generator is to produce accurate images of the target domain, whereas the discriminator distinguishes between the generated sample and the corresponding real sample. The feedback from the discriminator acts as an adaptive loss to guide the generator to do better in the transformation. In a nutshell, there is a min-max game going on between generator G and discriminator D. The objective function for a GAN is as follows:
\begin{equation}
    \begin{split}\label{eq1}
    V_{GAN}(G,D) &= E_{y\sim P_{data}(y)}[logD(y)] \\ &+ E_{z\sim P_{z}(z)}[log(1-D(G(z)))],
\end{split}
\end{equation}
where the generator, $G$, tries to minimize the optimization function and the discriminator, $D$, maximizes it.
An additional L2 or L1 loss term is added to the objective function to calculate the error between the input and output such that it penalizes the generator for creating dissimilar outputs. The final generator model is as follows:
\begin{align}\label{eq2}
    G_{optimal} = \underset{G}{min}\underset{D}{max} V_{GAN}(G,D)  +\lambda L_{L1}(y,G),
\end{align}
where $\lambda$ is the Lagrangian coefficient to control the relative strength of the reconstruction loss and $y$ is the ground truth fingerprint. The L1 distance is given by:
\begin{align}\label{eq3}
    L_{L1}(y,G) = \| y - G(z)\|_{1}.
\end{align}

In \cite{pix2pix2017}, Isola \textit{et al.} proposed a modification to the GAN model to train it in a constrained manner. In conventional GANs, the input to the network is not from a specific domain, but here, samples $x$ from the source domain are added as the input to the network. The discriminator is also constrained with the concatenated input of the target domain and the generated sample. After the modifications, the new objective function becomes:
\begin{equation}
    \begin{split}\label{eq4}
    V_{cGAN}(G,D) &= E_{x\sim P_{data}}[logD(x,y)] \\
    &+ E_{x\sim P_{data}}[log(1-D(x,G(x)))].
\end{split}
\end{equation}
\subsection{Multi-discriminator Deblurring}
We develop an additional modification to the cGAN model inspired by the multi-scale discriminator approach in \cite{8237891}. In addition to the main output of the generator that has the spatial size of $256\times 256$, we extract intermediate features maps at different layers of the generator e.g.,(spatial sizes $64\times64$ and $128\times128$) and force them to reconstruct the deblurred fingerprints. In this way, we provide additional supervision for the training of the generator. Hence, the generator will be guided more carefully toward estimating the deblurred fingerprint. To reduce the number of channels of the intermediate feature maps to an image, we use a $1\times1$ convolution. Afterward, we compare the deblurred images to the down-sampled version of the fingerprints. Our modified objective function for training the model is: 
\begin{equation}
\begin{split}\label{eqstackgan}
    V_{cGAN}&(G,D_1, D_2, D_3) = \\&E_{x\sim P_{data}}\big[\log\big(D_1(x_1,y_1)\big)\\ &+\log\big(D_2(x_2,y_2)\big)+\log\big(D_3(x_3,y_3)\big)\big] \\
    &+E_{x\sim P_{data}}\big[\log\big(1-D_1(x_1,G(x, 1))\big)\\
    &+\log\big(1-D_2(x_2,G(x, 2))\big)\\
    &+\log\big(1-D_3(x_3,G(x, 3))\big)\big],
    \end{split}
\end{equation}
where $\{x_1, x_2, x_3\}$ are the input fingerprints, $\{y_1, y_2, y_3\}$ are the ground truth fingerprints, and $G(x, 1), G(x, 2), G(x, 3)$ are the outputs of the generator at resolutions $64\times64$, $128\times 128$, and $256\times256$, respectively. $D_1, D_2,$ and $D_3$ are the three distinct discriminators that take the representations from the same generator as shown in Figure \ref{fig:network}. 
For the task of deblurring, the training data for source and target domains is available or can be generated synthetically. Therefore, we use the above model as the core network of our deblurring model. It accepts concatenated images of clean and blurry fingerprints and maps them to its corresponding deblurred image. However, unlike other types of data, fingerprints contain identity-specific information that needs to remain after deblurring. To this aim, we introduce additional sub-networks, such as a ridge extractor and a verifier module, that preserve the important information as well as improve the deblurring performance. The decomposition of the deblurring task using a multi-discriminatory approach along with ridge extractor and verifier showed promising results on the testing dataset. 
\subsection{Ridge Extraction Loss}
For retaining the fingerprint information, we include the ridge extraction loss into the generator loss function. Using the pre-trained weights of a ridge extractor, $G_{R}(\cdot)$, we calculate the L1 distance between the ridge pattern of the deblurred fingerprint and clean fingerprint. The loss term is given by:
\begin{align}\label{eq5}
    L_{L1}(G_{R}(y),G_{R}) = \| G_{R}(y) - G_{R}(G(z))\|_{1},
\end{align}
where y is the ground truth fingerprint and $G(z)$ is the deblurred fingerprint by the generator. Since the base model is a generative network, the synthesized fingerprints may look similar to the source, but, at the minutiae level, they do not match. Therefore, the ridge loss plays an important role in controlling the transformations made by the generator intact with the ridge pattern of  the fingerprint.
\begin{figure}[t]
\begin{center}
\includegraphics[width=0.8\linewidth]{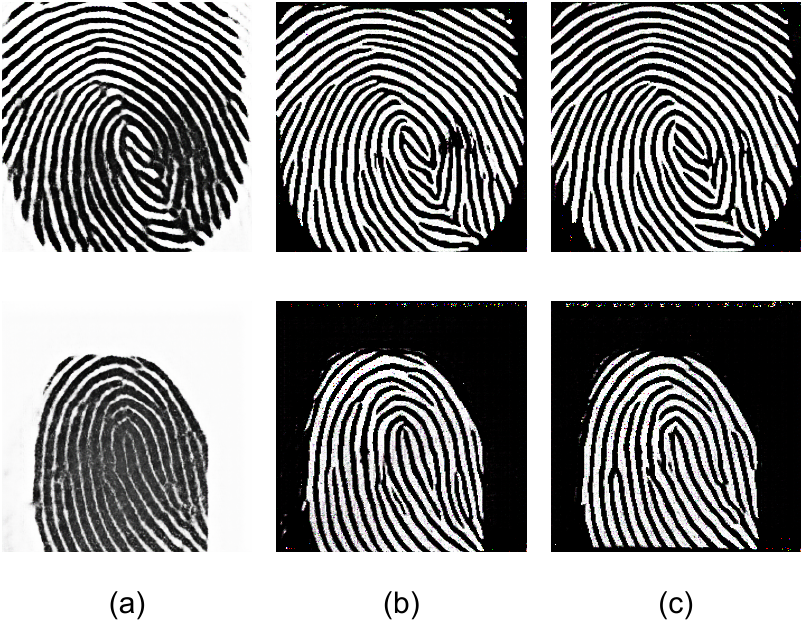}
\end{center}
   \caption{Ridge extraction results: Column (a) has deblurred fingerprints from the generator, (b) has ground truth ridge maps, and (c) has ridge maps generated by the ridge extractor.}
\label{fig:ridge_samples}
\end{figure}
\subsection{Verifier Loss}
In addition to the ridge extractor loss, we introduce another term to the final objective function which is the L2 verifier loss. Adding the L2 loss of all the intermediate representations of the verifier, balances the similarity of higher-level features such as finger shape, and lower-level features such as pores and minutiae between the generated fingerprint and the real fingerprint. It is given by:
\begin{align}\label{verification_cost}
    L_{L2}(S,G) = \sum_{i=1}^{n} \| S(y)_{i}  - S(G(z))_{i}\|^{2}_{2},
\end{align}
where n is the total number of residual blocks in the verifier module. $S(y)_i$, and $S(G(z))_i$ are the intermediate representations of the ground truth and deblurred fingerprint from the generator, respectively. They are extracted from $i^{th}$ residual block in the verifier network.
\subsection{Reconstruction Loss}
Due to the multi-discriminatory approach in the FDeblur-GAN, unlike a traditional GAN, we have three L1 reconstruction loss terms. It computes the distance between the intermediate deblurred fingerprints and the ground truth fingerprint such that the error from the intermediate layers is minimized. The reconstruction loss is given by: 
\begin{equation}
\begin{split}\label{reconstructionloss}
    L_{L1}(y,G) &= \| y_{1} - (G(z))_{1}\|_{1} + \| y_{2} - (G(z))_{2}\|_{1} \\&+\| y - (G(z))\|_{1},
\end{split}
\end{equation}
where $y_{1}$, $y_{2}$, and $y$ are ground truth fingerprints and $G(z)_{1}$, $G(z)_{2}$, and $G(z)$ are intermediate reconstructed fingerprints of size $64\times64$, $128\times128$, and $256\times256$, respectively.
\subsection{Objective Function}
The total objective function of the generator is the addition of all the cost terms. It also includes the loss computed from the intermediate deblurred fingerprints from the generator. Therefore, the updated cost function from Eq. \ref{eq2} is given by:
\begin{equation}
\begin{split}\label{eqfinalobj}
    G_{optimal} &= \underset{G}{min}\underset{D}{max} V_{cGAN}(G,D_1,D_2,D_3)  \\&+\lambda L_{L1}(y,G) \\ &+ \lambda_{R}L_{L1}(G_{R}(y),G_{R}) \\ &+ \lambda_{S}L_{L2}(S(y), S(G)),
    \end{split}
\end{equation}
where $L_{L1}(y,G)$ is the reconstruction loss given by Eq. \ref{reconstructionloss}. 
The scaling coefficients used for the loss terms are $\lambda$, $\lambda_{R}$, and $\lambda_{S}$. The latter two are the scaling coefficients for the ridge extraction and the verification loss respectively. They are set to $\lambda=100$ $\lambda_{R}=5$ and $\lambda_{S}=0.01$ using a grid search method.
\subsection{Network Architecture}
As stated above, the core of the proposed deblurring network is a cGAN model which deblurs the input image. During the training, the deblurred output at each intermediate layer is fed to its corresponding discriminator at that layer and the output of the last layer is fed to the ridge extractor and verifier. The intermediate features represent the characteristics of the low-resolution deblurred fingerprint images that allow the model to enhance the deblurring. The ridge extractor network is another cGAN model \cite{pix2pix2017} which takes concatenated inputs. The generator, $G_{R}$, is a pre-trained network, and is responsible for extracting ridge patterns from a given fingerprint image. It is trained on the dataset of the ridge patterns of the fingerprints. It minimizes the L1 distance between the generated ridge pattern and the ground truth ridge pattern.

The other sub-network is a Siamese verifier \cite{1467314} trained with contrastive loss \cite{1467314} using a ResNet-18 \cite{7780459} architecture. This network gets the same input as the last discriminator D3 in \ref{fig:network} and compares the two fingerprint samples. We also take the intermediate representations of the verifier and compute the L2 loss on each of them. Initially, the loss value of the intermediate features of each residual layer results in higher values depending on the similarity between the deblurred and ground truth fingerprint. However, as the generator tries to fool the discriminator and produces precisely similar fingerprints, this loss tends to reduce gradually. The verifier ensures the similarity of the input and output fingerprint images, whereas the ridge extractor tries to preserve the minutiae of the deblurred fingerprint. 

The generator and the ridge extractor both use a U-Net \cite{Ronneberger2015UNetCN}, \cite{pix2pix2017} architecture. The U-Net architecture has skip connections from the encoder to the decoder which enhances the quality of the decoder. Figure \ref{fig:network} shows the details of our FDeblur-GAN architecture. The discriminator network is a three-layer PatchGAN discriminator \cite{pix2pix2017}. It classifies if the image is real or fake on every 70x70 patch of the output of the generator. 
%--------------------------------------------------------------------------
\section{Experiments}
In this section, we cover the training methodology and the different experiments that we performed for testing the FDeblur-GAN model accompanied by the results. Further, we explore the effects of deblurring on the identification process.
\subsection{Training}
\begin{figure}[t]
\begin{center}
\includegraphics[width=1.0\linewidth]{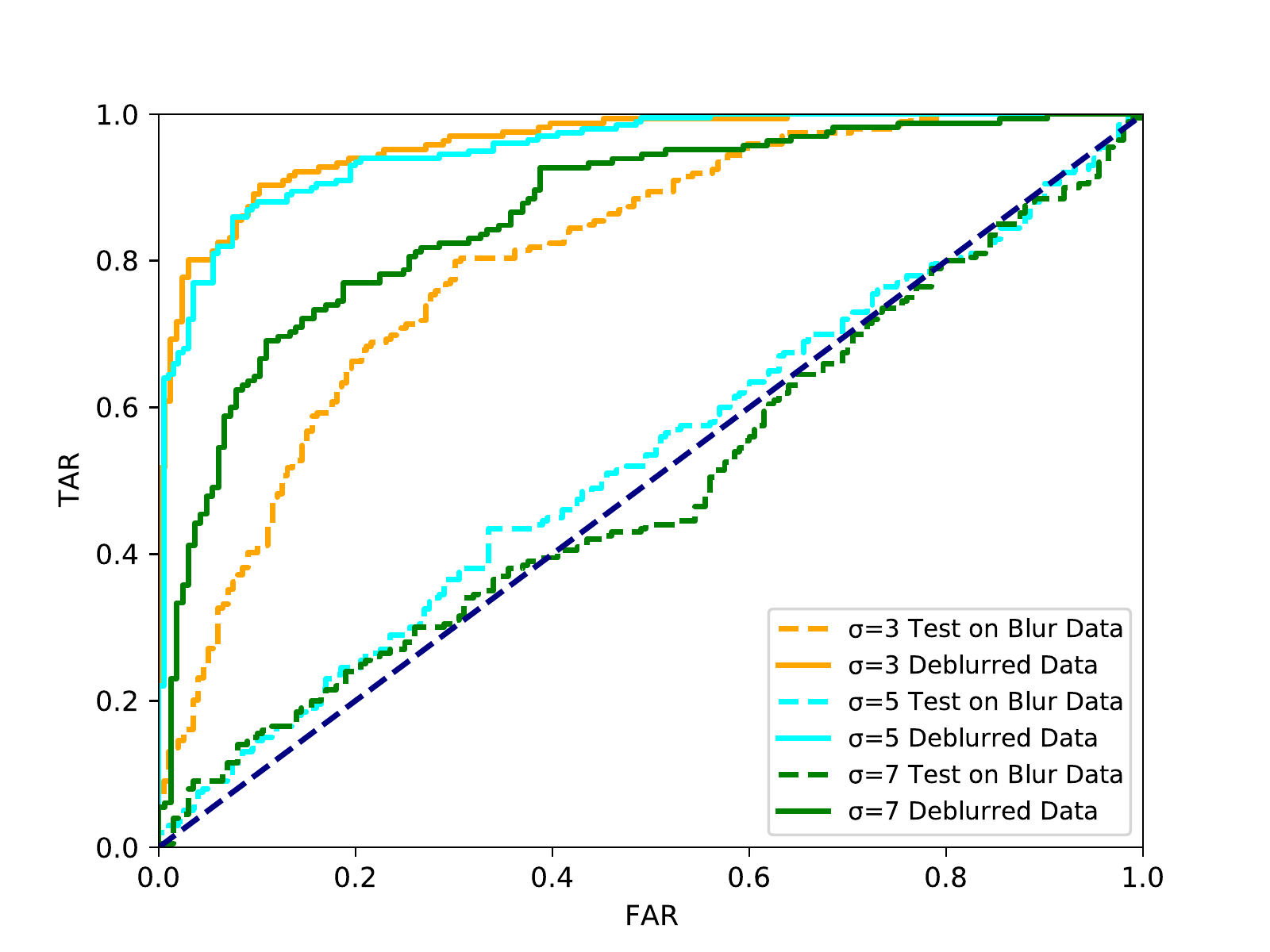}
\end{center}
   \caption{Deblurring ROC curve from tests using blur and deblurred data. Here, TAR and FAR are True Acceptance Rate and False Acceptance Rate, respectively. Detailed scores are given in Table \ref{tab:results}.}
\label{fig:blur_deblur_roc}
\end{figure}
We trained the proposed network for 50 epochs which took around 9 hours with two Nvidia Titan X GPUs. The ridge extractor is trained for 100 epochs. In both networks, the Adam optimizer \cite{kingma2017adam} is used as the optimizer with initial learning rate of $2\times10^{-4}$. The momentum parameters used are: $\beta_{1}=0.5$ and $\beta_{2}=0.999$.
\subsection{Dataset and Preprocessing}
For training and testing, we used the newly collected Multimodal Biometric Dataset\footnote{Dataset available on request to authors.} which consists of 15,860 fingerprint images of 250 subjects. Out of these, 13,734 images were used for training, and others were used for validation and testing purposes. During the experiment, we noticed that the blurred fingerprint data is not readily available, and therefore, we created a synthetic dataset using a Gaussian blur with different kernel sizes. The kernel size for the Gaussian blur function is given by: $k= 6\sigma -1$, where $k$ is the input fingerprint image and $\sigma$ values are among 3, 5, or 7. 

In the training of the ridge extractor sub-network, we used the Gabor filter on the fingerprint images to extract the ridge information. These images were used as the ground truth for the discriminator to distinguish between the generated ridge and ground truth ridge pattern.

Preparing the dataset for the deblurring network included preprocessing on clean fingerprints. Collected data contains noise, misaligned finger positions, etc. To eliminate such problems we first perform finger ridge segmentation using Gabor responses \cite{709565}. Then, we extract the fingerprint core using the directional histogram of the directional image of the fingerprint \cite{SRINIVASAN1992139}. After extracting the core point, we crop a 256x256 image around the core and use that as the input to the FDeblur-GAN model. 
\subsection{Evaluation}
\begin{figure}[t]
\begin{center}
\includegraphics[width=1.0\linewidth]{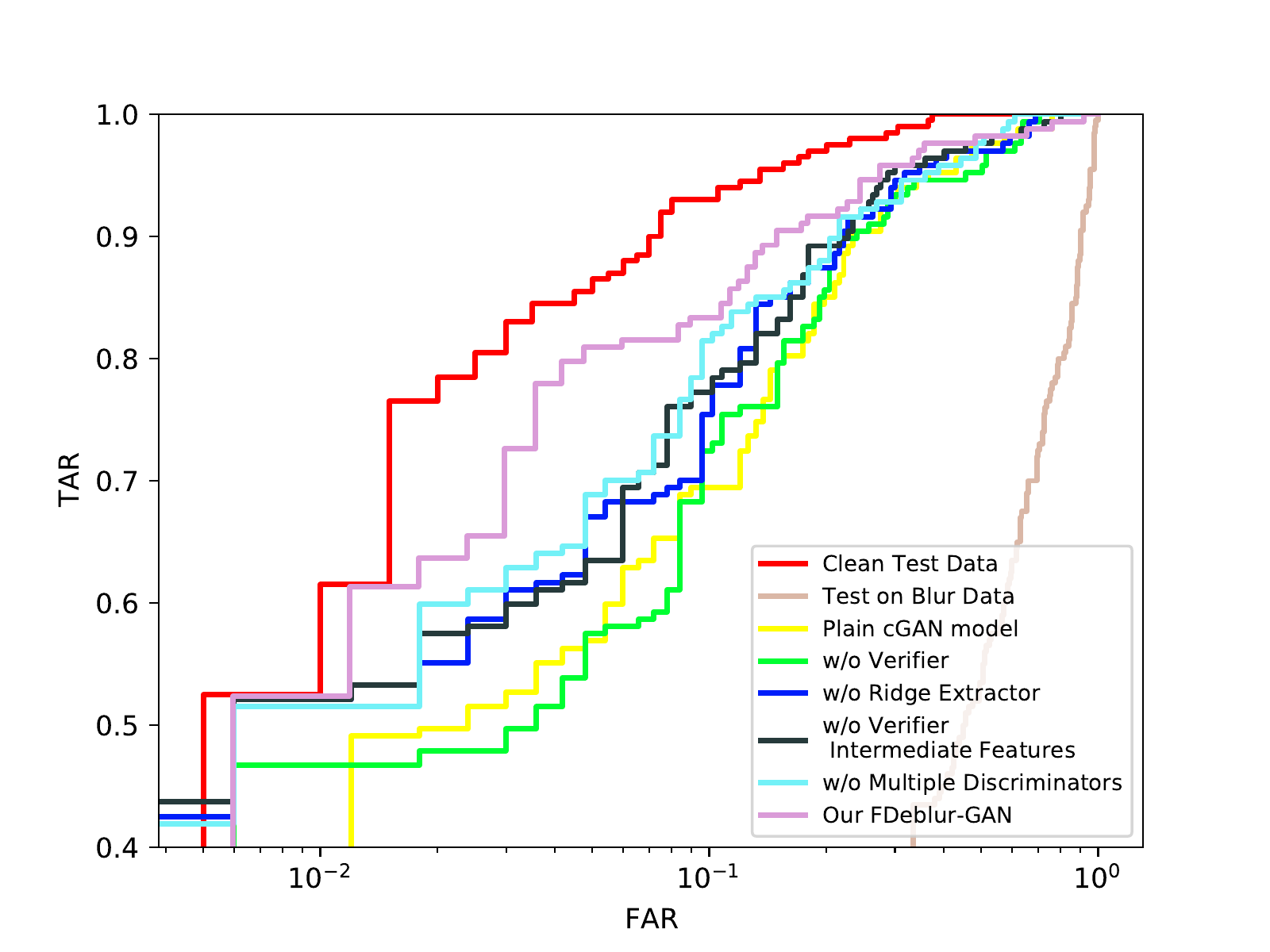}
\end{center}
   \caption{Log scaled ROC curves of different experiments performed during the ablation study. Here, TAR and FAR are True Acceptance Rate and False Acceptance Rate, respectively. The AUC and EER details are in Table \ref{tab:ablation_results}.}
\label{fig:ablation_roc}
\end{figure}

\begin{figure}[t]
\begin{center}
\includegraphics[width=1.0\linewidth]{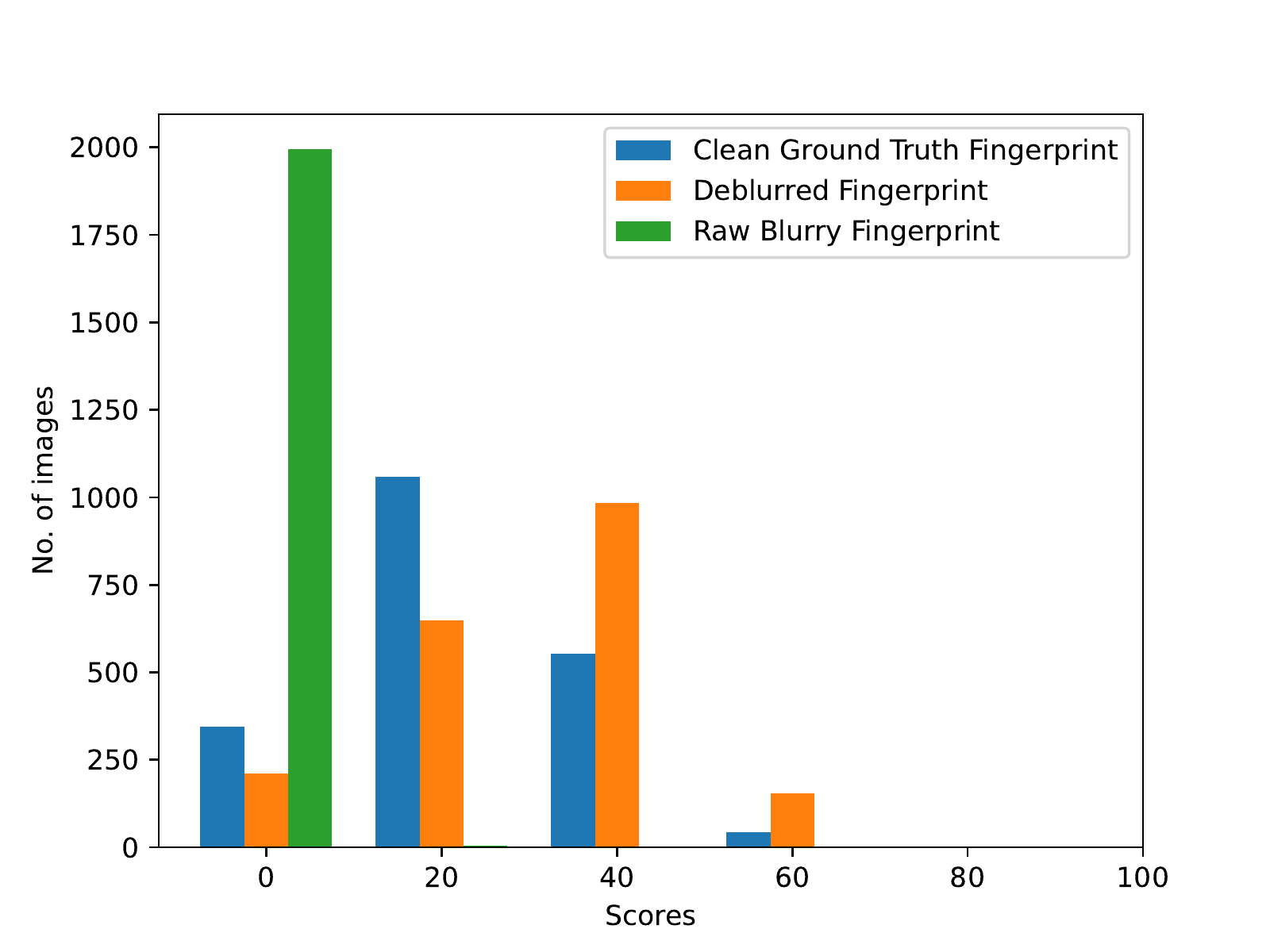}
\end{center}
   \caption{Quality score assesment of the ground truth, deblurred, and blurred fingerprints.}
\label{fig:quality_scores}
\end{figure}

Figure \ref{fig:deblurred_samples} and Figure \ref{fig:ridge_samples} depicts the deblurred fingerprints and ridge extraction results, respectively. For evaluating the performance, matching experiments were conducted on the original, blur and deblurred images using a Siamese-like fingerprint verifier \cite{1467314}. As shown in Figure \ref{fig:ablation_roc}, the recognition performance of the proposed FDeblur-GAN model follows closely the curve of the clean fingerprint curve. Figure \ref{fig:blur_deblur_roc} shows the recognition performance on the blurred data for different $\sigma$ values. Our model achieves a high accuracy in the deblurring task. Figure \ref{fig:quality_scores} represents the quality scores calculated using the NFIQ2 tool provided by NBIS software \cite{nbis}. The NFIQ2 tool provides quality scores from 1 to 100, where 1 is the worst and 100 is considered the best. Our model preserves and in some cases improves the quality of the ground truth fingerprints during the deblurring process.
During the experimentation, we observed that if the $\sigma$ value is lower than three, the fingerprints can be verified without deblurring. On the other hand, for $\sigma$ value greater than seven, the fingerprints lose necessary information and become intractable. Blurred samples along with their clean samples are shown in Figure \ref{fig:deblurred_samples}.
%------------------------------------------------------------------------
\begin{figure}
\begin{center}
\includegraphics[width=0.8\linewidth]{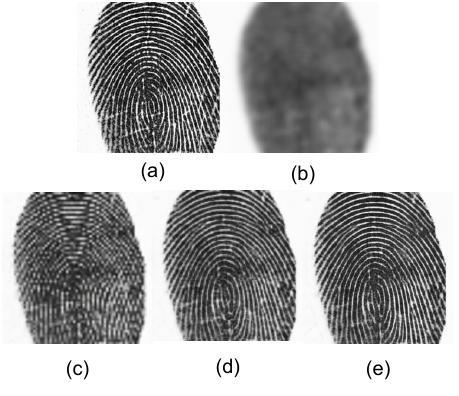}
\end{center}
   \caption{Intermediate representations along-with input and ground truth image. (a) the ground truth fingerprint, (b) the blurred input, (c) the first intermediate representation of size 64x64, (d) the second intermediate representation of size 128x128, and (e) the deblurred output of size 256x256.}
\label{fig:intermediate_results}
\end{figure}
\begin{table}
    \centering
    \setlength{\tabcolsep}{9pt} % Default value: 6pt
    \caption{It presents the results of the FDeblur-GAN model trained on images blurred using different $\sigma$ values. Here, $\sigma$ is the standard deviation of the Gaussian blur applied to the fingerprint images.}
    \begin{tabular}{c  c  c  c}
    $\sigma$ & Data & EER & AUC \\
    \hline
    \multirow{2}{*}{$3$} & w/o deblurring & $0.2714$ & $0.8054$ \\
    \cline{2-4}
    & w/ deblurring & $\textbf{0.1024}$ & $\textbf{0.9607}$\\
    \hline
    \multirow{2}{*}{$5$} & w/o deblurring & $0.4800$ & $0.5279$\\
    \cline{2-4}
    & w/ deblurring & $0.1200$ & $0.9518$\\
    \hline
    \multirow{2}{*}{$7$} & w/o deblurring & $0.5450$ & $0.4952$\\
    \cline{2-4}
    & w/ deblurring & $0.2242$ & $0.8642$
    \end{tabular}
    \label{tab:results}
\end{table}
Table \ref{tab:results} shows the results of the verification experiment performed on the blurred data and the same experiment performed on the deblurred data. The lower EER demonstrates the ability of the model to deblur the fingerprints and preserve the necessary identity information.
\begin{table}
    \centering
    \setlength{\tabcolsep}{6pt} % Default value: 6pt
    \caption{\label{tab:ablation_results} Results of each model tested for the ablation study. Note that these models are trained with $\sigma=5$.}
    \begin{tabular}{ L c c}
    Model & EER & AUC \\
    \hline
    Plain cGAN model & $0.1796$ & $0.9090$ \\
    \hline
    Deblurring without verifier & $0.1737$ & $0.9091$ \\
    \hline
    Deblurring without ridge extractor & $0.1497$ & $0.9261$ \\
    \hline
   Deblurring without verifier intermediate features & $0.1617$ & $0.9287$ \\
   \hline
    Deblurring without multiple discriminators & $0.1497$ & $0.9321$ \\
    \hline
    Proposed deblurring model & $\textbf{0.1200}$ & $\textbf{0.9518}$ \\
    \end{tabular}
\end{table}
\section{Ablation Study}
In the ablation study, we present experiments by excluding one of the sub-networks from the proposed model and study its effect on the overall deblurring as well as identification performance. 
\subsection{Deblurring with Plain Model}
First, to determine the performance of the cGAN \cite{pix2pix2017}, we trained the model without any additional sub-networks. With the advent of image-to-image translation \cite{pix2pix2017}, the deblurring task is primitive for the network, but preserving the necessary information is challenging. Table \ref{tab:ablation_results} shows the results of the plain deblurring model. We use this performance as a baseline for further experiments. As illustrated, the plain cGAN model performs really well at transforming input fingerprints from the blurred domain to the deblurred domain. It generated clear deblurred images, but the images had incorrect ridge patterns. Such discrepancy may alter the ID of the fingerprint affecting recognition performance. Based on this observation, we constrained the generator with a ridge extractor and a verifier sub-network.
\subsection{Evaluation without Ridge extractor or Verifier}
Further, we tested the proposed FDeblur-GAN network without the ridge extractor and removed the L1 loss term in Eq. \ref{eq5} from the overall network objective function in Eq. \ref{eqfinalobj}. In the case of fingerprint deblurring, the best way to evaluate the model performance is by monitoring the identification performance. Figure \ref{fig:ablation_roc} and Table \ref{tab:ablation_results} shows the degraded performance without the ridge extractor. The ridge extractor is one of the constraints on the generator which restricts it to deblur the images while keeping the ridge pattern similar. Therefore, the performance is significantly lower than the best score.  

The verifier plays a vital role in the training. It enforces the generator to preserve the ID information from the blurred fingerprint. As the verification cost term in Eq. \ref{verification_cost} is removed from the final objective function in Eq. \ref{eqfinalobj}, it resulted in a 4.5\% drop in the accuracy of the identification task.  
Finally, we tested the network by excluding the loss of the intermediate features of the verifier. As shown in Table \ref{tab:ablation_results}, due to the generalized characteristics in the intermediate features, the network performs better in identifying the fingerprints.  According to Figure \ref{fig:ablation_roc}, without the intermediate feature loss, the accuracy in the lower FAR region is low. However, adding the loss term improves the accuracy in the specifically lower FAR region, which is widely considered in real-world applications.
\subsection{Effect of multi-discriminatory approach}
As discussed in Section \ref{sec:cgan}, we developed a multi-discriminator model to enhance the performance of deblurring. Our results in Table \ref{tab:ablation_results} and Figure \ref{fig:ablation_roc} demonstrate that guiding the generator by forcing the intermediate features to mimic the coarse structure of the ground truth images indeed improves the performance of deblurring. Particularly, based on Table \ref{tab:ablation_results}, employing multiple discriminators improves EER and AUC  by $19.8\%$ and  $2.1\%$, respectively.

%------------------------------------------------------------------------
\section{Conclusion}
We proposed a method of deblurring fingerprints using the conditional GAN and utilizing the multi-layer features of the stack GAN \cite{8237891} along with two sub-networks. We thoroughly tested our model with different parameters to improve the performance of the network. In addition to this, our ablation study illustrates improvement achieved by different loss terms in the network. We successfully deblurred fingerprints while retaining their quality. FDeblur-GAN is a novel approach that has various real-world applications in areas such as crime scene investigation and forensic sciences.  
{\small
\bibliographystyle{ieee}
\bibliography{egbib}
}

\end{document}